\documentclass{article}

\usepackage[preprint]{neurips_2023}

\usepackage[utf8]{inputenc} 
\usepackage[T1]{fontenc}    
\usepackage{hyperref}       
\usepackage{url}            
\usepackage{booktabs}       
\usepackage{amsfonts}       
\usepackage{nicefrac}       
\usepackage{microtype}      
\usepackage{xcolor}         
\usepackage{multirow}
\usepackage{graphicx}
\usepackage{subfigure}
\usepackage{subcaption}
\usepackage{amsmath}
\usepackage{mathrsfs}
\usepackage{amsthm}
\usepackage{algorithm}
\usepackage{algorithmic}

\newtheorem{definition}{Definition}
\newtheorem{theorem}{Theorem}
\newtheorem{lemma}{Lemma}

\title{Equitable Multi-Task Learning}

\author{%
  Jun Yuan \\
  Huawei Noah’s Ark Lab\\
  \texttt{yuanjun25@huawei.com} \\
   \And
   Rui Zhang \\
   www.ruizhang.info \\
   \texttt{rayteam@yeah.net}
}

\begin{document}

\maketitle

\begin{abstract}
  Multi-task learning (MTL) has achieved great success in various research domains, such as CV, NLP and IR etc. Due to the complex and competing task correlation, na\"ive training all tasks may lead to inequitable learning, \textit{i.e.} some tasks are learned well while others are overlooked. Multi-task optimization (MTO) aims to improve all tasks at same time, but conventional methods often perform poor when tasks with large loss scale or gradient norm magnitude difference. To solve the issue, we in-depth investigate the equity problem for MTL and find that regularizing relative contribution of different tasks (\textit{i.e.} value of task-specific loss divides its raw gradient norm) in updating shared parameter can improve generalization performance of MTL. Based on our theoretical analysis, we propose a novel multi-task optimization method, named \textit{EMTL}, to achieve equitable MTL. Specifically, we efficiently add variance regularization to make different tasks' relative contribution closer. Extensive experiments have been conduct to evaluate EMTL, our method stably outperforms state-of-the-art methods on the public benchmark datasets of two different research domains. Furthermore, offline and online A/B test on multi-task recommendation are conducted too. EMTL improves multi-task recommendation significantly, demonstrating the superiority and practicability of our method in industrial landscape. \footnote{We will make our source code public after paper accepted.}
\end{abstract}

\section{Introduction}

Multi-task learning (MTL)~\cite{Caruana97,ruder2017overview} simultaneously learns multiple tasks via a joint model and has achieved great success in various applications, such as text classification~\cite{mao2021banditmtl}, domain adaption~\cite{Fangrui21pareto} and Computer Vision~\cite{liu2021imtl} etc. However, due to the complex and competing task correlation, na\"ive training all tasks may lead to inequitable learning, \textit{i.e} some tasks are learned well while others are overlooked. Several methods have been proposed to solve inequitable learning problem in optimization, which are called Multi-task optimization (MTO). 

Previous studies~\cite{liu2021imtl,navon2022nashmtl,9893398} have shown that large loss scale or gradient norm magnitude difference of the tasks can be one primary obstacle against improving all tasks. Moreover, unbalanced loss or gradient norm is very common in practice~\cite{DBLP:journals/corr/ZhangY17aa}. Some loss scale free methods have been proposed~\cite{navon2022nashmtl,liu2021imtl} recently. However, conventional MTL methods only consider the equity between various tasks in an over-simplified manner. For example, IMTL~\cite{liu2021imtl} straightforwardly forces the aggregated gradient to have equal projections onto individual tasks' raw gradient, or scales all task-specific losses to be the same. Such a naive method does  guarantee convergence~\cite{navon2022nashmtl}. 

In this work, we find that regularizing relative contribution in updating shared parameters of different tasks (\textit{i.e.} $\mathcal{L}_{t}/||\boldsymbol{g}_{t}||$)  can help MTL models get smaller generalization error, rather than simply constraining absolute contribution (\textit{i.e,} $||\boldsymbol{g}_{t}||$) like IMTL. We prove that the expected difference between empirical relative contribution and generalization relative contribution is upper bounded by the variance of empirical relative contribution. We call the method that can make relative contribution of all tasks to be the same as equitable MTL. To achieve equitable MTL, we propose a novel method \textit{EMTL} to efficiently add regularization on the variance of empirical relative contribution. EMTL is robust against large loss scale or gradient norm difference by balancing gradient and loss at same time. 



\begin{figure}[htp]

    \includegraphics[width=\textwidth]{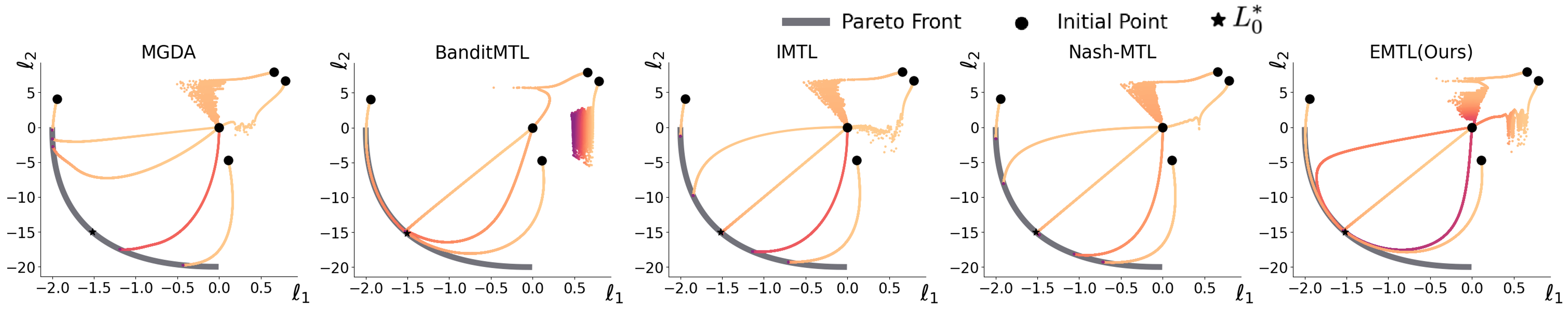}
    \caption{\textit{Toy experiments}. Experiments start from 5 different initialization points (black dots •) and their trajectories. The trajectories are colored from orange to purple. Losses have a large difference in scale. See Appendix for details. Our method, EMTL, is robust to changes in loss scale and can converge to optimal parameters starting from all initialization points}\label{toy}

\end{figure}

We extensively evaluate our proposed EMTL on public benchmark datasets on two different research domains, Computer Vision (CV) and Recommendation. We also conduct experiments on industrial dataset, which is collected from commercial recommendation system. The industrial dataset includes both classification and regression tasks. We further conduct online AB test by combining our method with MTL model. The results demonstrate that EMTL not only outperforms SOTA methods, but also has great practical value in industrial landscape. 

In summary, our principal contributions are listed as follows:
\begin{itemize}
    \item We propose the thought to achieve equitable MTL and further improve MTL by regularizing relative contribution of tasks in updating shared parameters. Moreover, we theoretically verifies that variance regularization of relative contribution of all tasks can improve the generalization performance of MTL.
    \item Based on our analysis, we propose a novel method \textit{EMTL} to realize equitable MTL, which can add variance regularization to conventional MTL objective efficiently. Our method can balance gradient and loss simultaneously and is robust against to loss scale or gradient norm magnitude changes.
    \item We extensively evaluate the effectiveness of our EMTL on two different research domains, along with industrial dataset. The experimental results demonstrate the superiority of our method and it is more robust against loss scale changes compare to SOTA methods.
\end{itemize}





\section{Related Work}

MTL simultaneously learns the tasks by minimizing their empirical losses together, thus some specific tasks are learned well while others are overlooked, due to the different \textit{loss scales} or \textit{gradient norm magnitudes} of tasks and the \textit{mutual competition} among tasks. Multi-task optimization is to address the problem of simultaneously minimizing multiple objectives. Multi-task optimization often try to find loss weights to be multiplied on the raw losses for model optimization. Generally, MTO methods can be divided into two categories~\cite{DBLP:journals/corr/ZhangY17aa}, Gadient Balancing Method (GBM) and Loss Balancing Method (LBM).

\textbf{GBM}s try to alleviate mutual competition among tasks by combining per-task gradients of shared parameters into a joint update direction using a particular heuristic. MGDA~\cite{mgda} casts multi-task learning as multi-object optimization and finds the minimum-norm point in the convex hull composed by the gradients of multiple tasks. MGDA-UB~\cite{DBLP:conf/nips/SenerK18} propose an approximation of the original optimization problem of MGDA to improve efficiency. PCGrad~\cite{Yu20pcgrad} avoids interferences between tasks by projecting the gradient of one task onto the normal plane of the other. CAGrad~\cite{liu2021cagrad} optimizes for the average loss while explicitly controlling the minimum decrease rate across tasks. AdaTask~\cite{yang22adatask} uses task-specific accumulative gradients when adjusting the learning rate of each parameter. GBMs can evenly learn task-shared parameters while ignoring task-specific ones, making them performance poor when loss scale of tasks are huge difference. \textbf{LBM}s use delicate heuristic to integrate loss of various tasks based on some assumptions. Uncertainty weighting~\cite{Kendall2018uncertainty} models the loss weights as data-agnostic task-dependent homoscedastic uncertainty. Then loss weighting is derived from maximum likelihood estimation. GradNorm~\cite{chen2018gradnorm} learns the loss weights to enforce the norm of the scaled gradient for each task to be close. BanditMTL~\cite{mao2021banditmtl} adds variance of task losses to regularize optimization, but it focus on the task with large loss scale. So it needs to adopt some tricks to balance loss scale. LBMs often have higher efficiency than GBM, and some LBM, such as GradNorm, can prevent MTL from being biased in favor of tasks with large loss scales. However, they may not be adopted when their assumptions do not hold.

IMTL~\cite{liu2021imtl} and NashMTL~\cite{navon2022nashmtl} are loss scale free GBM. IMTL-G makes the aggregated gradient has equal projections onto individual tasks, and IMTL-L scale all task-specific loss to be same. NashMTL uses Nash bargaining solution making tasks negotiate to reach an agreement on a joint direction of parameter update in the ball of fixed radius centered around zero. The strict restriction of above two methods makes them can not improve MTL model consistently. Detailed discussion about EMTL and IMTL please refer to Section~\ref{sec:discussion}.

As shown in Figure~\ref{toy}, we conduct a toy experiment to intuitively understand difference of several SOTA methods, and $\mathcal{L}^{*}_{0}$ is the best average loss of $\mathcal{L}_{1}$ and $\mathcal{L}_{2}$. BanditMTL can not make all runs converge to Pareto set, because it is sensitive to large loss scale. MGDA focuses on the task with smaller gradient norm. Though it can converge to Pareto front, but it converges to the top left or bottom right, making one task performance poor. As to NashMTL and IMTL, we find it can not converge to the optimal solution $\mathcal{L}^{*}_{0}$, because of their strict limitations. EMTL is robust to changes in loss scale and can converage to optimal solution stablely.

\section{Theoretical Analysis of Equity for MTL}\label{sec:analysis}

\subsection{Definition of Equitable MTL}

Before defining equitable MTL, we first define a metric to measure relative contribution of tasks in updating shared parameters.

The raw gradient's norm of task \textit{t} on shared parameters can be seen as the absolute contribution in updating shared parameters. So we define the following metric to measure the relative contribution of task \textit{t} in updating shared parameters.
\textbf{Relative rate of loss and gradient norm} of task \textit{t} is defined as follow:
\begin{equation}
    Rr_{t} = \frac{\mathcal{L}_{t}(\theta)}{||\boldsymbol{g}_{t}||}
\end{equation} $\mathbf{g}_t$ is the raw gradient of current update step on shared parameters of task \textit{t}. To simplify, we call relative rate of loss and gradient norm as relative rate in following.



\textbf{Equitable MTL}:
Based on the above metric, we define equitable MTL as follow: Different tasks have equal relative rate of loss and gradient norm on shared parameters. 

Then, the mini-batch equitable MTL can be formally defined as following formulation:
\begin{equation}
    \frac{\mathcal{L}_{1}(\theta)}{||\boldsymbol{g}_{1}||}=\frac{\mathcal{L}_{t}(\theta)}{||\boldsymbol{g}_{t}||}=...=\frac{\mathcal{L}_{T}(\theta)}{||\boldsymbol{g}_{T}||}
\end{equation} T is the number of tasks that trained simultaneously. 
Intuitively, equable multi-task learning means the contribution of each task (i.e. $||\boldsymbol{g}_{t}||$) should be proportional to its loss. The theoretical analysis of relative rate please refer to the next section and Appendix.

\subsection{Theoretical Analysis}\label{sec:theo_analysis}
In this section, we give the generalization error bound of relative rate in Theorem~\ref{theorem_1}. The theorem indicates that variance regularization can improve generalization ability of MTL model. To derive the error bound, we follow the analysis procedure of BanditMTL~\cite{mao2021banditmtl}. 1) We first give the upper bound for the representativeness of training set. 2) We give  the Rademacher complexity-based generalization error bound based on this representativeness bound. 3) We derive the variance-based generalization error bound based on the Rademacher complexity-based bound.

For convenience of writing, we define some important terms first. The empirical loss of task \textit{t} is defined as $\hat{\mathcal{L}}(\theta) = \frac{1}{N} \sum_{n=1}^{N} \mathcal{L}_{t}(\theta, x_{t}, y_{t})$. The generalization loss of task \textit{t} is defined as $\mathcal{L}_{t}(\theta) = \mathbb{E}_{(x_{t},y_{t}) \sim \mathcal{D}_{t}}\mathcal{L}_{t}(\theta, x_{t},y_{t})$, where $\mathcal{D}_{t}$ is the data distribution of task \textit{t}. Training set is denoted as $S$, and $N = |S|$ represents the training data number.

\begin{definition}
(Relative Rate Variance). For an MTL model with parameter $\theta$, let $\hat{\mathcal{L}}_{t}(\theta)$ denote the empirical loss of task $t \in [1,T]$, and the raw gradient of task t is $\hat{\boldsymbol{g}}_{t}$. Then the relative rate variance is defined as 
\begin{equation}
    Var(\theta) = \sqrt{\frac{1}{T} \sum_{t=1}^{T} (\hat{\mathcal{L}}_{t}(\theta)/||\hat{\boldsymbol{g}}_{t}||-\frac{1}{T}\sum_{t=1}^{T}\hat{\mathcal{L}}_{t}(\theta)/||\hat{\boldsymbol{g}}_{t}||)^{2}}
\end{equation}
\end{definition}

We first give the upper bound for the representativeness of training set $S$ as Lemma~\ref{lemma_1}.
\begin{lemma}\label{lemma_1}
Assume $\forall (x_{t}^{n}, y_{t}^{n}), (x_{t}^{m}, y_{t}^{m}): |\mathcal{L}_{t}(\theta, x_{t}^{n}, y_{t}^{n})/||\boldsymbol{g}^{n}_{t}||-\mathcal{L}_{t}(\theta, x_{t}^{m}, y_{t}^{m})/||\boldsymbol{g}^{m}_{t}||| < c$. Let 
\begin{equation}
    Rep(\mathcal{H}, S) = \sup_{\theta \in \mathcal{H}} \frac{1}{T} \sum_{t=1}^{T} (\mathcal{L}_{t}(\theta)/||\boldsymbol{g}_{t}|| - \hat{\mathcal{L}}_{t}(\theta)/||\hat{\boldsymbol{g}}_{t}||),
\end{equation} then $\forall \delta \in [0, 1]$, with probability of at least $1 - \delta$:
\begin{equation}
    Rep(\mathcal{H}, S) \leq \mathbb{E}_{S} Rep(\mathcal{H}, S) + c\sqrt{\frac{2\log(2/\delta)}{TN}}
\end{equation}
\end{lemma}
\begin{proof}
The proof is provided in Appendix
\end{proof}

Lemma~\ref{lemma_1} yields that the representativeness of relative rate on $S$, $Rep(\mathcal{H}, S)$ , is bounded by its expected value. Combining Lemma~\ref{lemma_1} and standard symmetrization argument, we give the Rademacher complexity-based generalization error bound for MTL as Lemma~\ref{lemma_2}
\begin{lemma}\label{lemma_2}
Let $\sigma = \{\{\sigma_{t}^{n}\}_{t=1}^{T}\}_{n=1}^{N}$ be a sequence of binary random variables. Each $\sigma_{t}^{n} = \pm 1$ is independent with probability 0.5, and denote $R(l \circ \mathcal{H} \circ S) = \mathbb{E}_{\sigma} \sup_{\theta}(\frac{1}{TN} \sum_{t=1}^{T} \sum_{n=1}^{N} \sigma_{t}^{n} \mathcal{L}_{t}(\theta, x_{t}^{n}, y_{t}^{n})/||\boldsymbol{g}_{t}^{n}||)$. Then, for $\forall \delta \in [0,1]$, with probability of at least $1-\delta$:
\begin{equation}
    \frac{1}{T} \sum_{t=1}^{T} (\frac{\mathcal{L}_{t}(\theta)}{||\boldsymbol{g}_{t}||} - \frac{\hat{\mathcal{L}}_{t}(\theta)}{||\hat{\boldsymbol{g}}_{t}||}) \leq 2 R(l \circ \mathcal{H} \circ S) + 3c\sqrt{\frac{2\log(4/\delta)}{TN}}
\end{equation}
\end{lemma}
\begin{proof}
The proof is provided in Appendix
\end{proof}

Based on Lemma~\ref{lemma_2}, independence of random variables and property of convex function, we give the generalization error bound of relative rate variance as Theorem~\ref{theorem_1}.

\begin{theorem}\label{theorem_1}
Consider an MTL problem with T tasks, training set S and hypothesis class $\mathcal{H}$, and let $\hat{\mathcal{L}}_{t}(\theta)$ denote the empirical loss of task $t \in [1,T]$. If $\forall (x_{t}^{n}, y_{t}^{n}), (x_{t}^{m}, y_{t}^{m}): |\mathcal{L}_{t}(\theta, x_{t}^{n}, y_{t}^{n})/||\boldsymbol{g}^{n}_{t}||-\mathcal{L}_{t}(\theta, x_{t}^{m}, y_{t}^{m})/||\boldsymbol{g}^{m}_{t}||| < c$. Then, for $\forall \delta \in [0, 1]$ with probability of at least $1-\delta$, for all $\theta \in \mathcal{H}$,
\begin{equation}
    \frac{1}{T} \sum_{t=1}^{T} (\frac{\mathcal{L}_{t}(\theta)}{||\boldsymbol{g}_{t}||} - \frac{\hat{\mathcal{L}}_{t}(\theta)}{||\hat{\boldsymbol{g}_{t}}||}) \leq 2 \sup_{\theta \in \mathcal{H}} Var(\frac{\hat{\mathcal{L}}_{t}(\theta)}{||\hat{\boldsymbol{g}_{t}}||}) + 3c\sqrt{\frac{2\log(4/\delta)}{TN}}
\end{equation}
\end{theorem}
\begin{proof}
The proof is provided in Appendix
\end{proof}

Theorem~\ref{theorem_1} states that generalization relative rate is upper bounded by the empirical relative rate variance. Thus, regularizing the relative rate variance can improve the generalization performance of MTL. Note that above analysis is based on updating parameters with whole training set, but we calculate loss, gradient and variance on mini-batch data in practice. Besides, equitable MTL can not prevent task conflict. To leverage Theorem 1 and solve above two problems, we propose our EMTL in the next section.




\section{The Proposed Approach}



\subsection{EMTL}\label{sec:method}

In this section, we describe our EMTL in detail. According to the definition and analysis on equity for MTL, the variance of relative rate will increase when tasks are optimized inequitably. Intuitively, we can add variance regularization of relative rate during training, and the straightforward choice of objective function can be formulated as Eq~\ref{eq:obj1}:
\begin{equation}
    \min_{\theta} \{\frac{1}{T} \sum_{t=1}^{T} \mathcal{L}_{t}(\theta) + \rho \mathrm{Var}(Rr) \}\label{eq:obj1}
\end{equation} where $\mathrm{Var}(Rr) = \sqrt{\frac{1}{T} \sum_{t=1}^{T} (Rr_{t} - \frac{1}{T} \sum_{t=1}^{T} Rr_{t})^{2}}$ is the empirical variance between task-specific relative rate, and $\rho$ is the hyperparameter to control the trade-off between the mean empirical loss the relative rate variance. However, formulation~\ref{eq:obj1} is generally non-convex and associated NP-hardness. 

Actually, to achieve equitable MTL and improve all tasks, we need to solve two sub-problems: 1) How to avoid conflict between tasks? 2) How to add variance regularization efficiently?  


Firstly, simply sum all task losses commonly leading to some tasks suffer from performance degeneration, due to the gradient conflict. We can adopt existed GBM, like MGDA, to alleviate this problem. After getting the weights generated by a GBM of every task $\boldsymbol{\alpha} = (\alpha_{1},...,\alpha_{T})$, we try to regularize the relative rate of weighted loss. That is to say, we let $$\tilde{Rr}_{t} = \alpha_{t} Rr_{t}=\frac{\alpha_{t}\mathcal{L}_{t}}{||\boldsymbol{g}_{t}||}$$ Then we regularize $\tilde{Rr}_{t}$ of all tasks.

Secondly, we propose an efficient method to add variance regularization on relative rate of MTL objective. Eq~\ref{eq:obj1} is still non-convex and associated NP-hardness, even $\mathcal{L}_{t}(\theta)$ is convex in $\theta$, yielding computationally intractable problems. According to~\cite{Ben-TalHWMR13,BertsimasGK18a,mao2021banditmtl}, we select a convex surrogate based on its equivalent formulation.
\begin{equation}
    \sup_{\mathbf{p} \in  \mathcal{P}_{\rho, \mathbf{T}}} \frac{1}{T} \sum_{t=1}^{T} p_{t} \tilde{Rr}_{t} = \frac{1}{T} \sum_{t=1}^{T} \tilde{Rr}_{t} + \rho \sqrt{\mathrm{Var}(\tilde{Rr}_{t})} + o(T^{\frac{1}{2}}) \label{eq:surrogate1}
\end{equation} where $\mathcal{P}_{\rho, \mathbf{T}} := \{ \mathbf{p} \in \mathbb{R}^{T}:\ \sum_{t=1}^{T} p_{t} = 1,\ p_{t} \geq 0,\ \sum_{t=1}^{T} p_{t}\log(Tp_{t}) \leq \sqrt{\rho} \}$ and $\mathbf{p}$ is the weights vector sampled in the area $\mathcal{P}_{\rho, \mathbf{T}}$.   

By further analysis Eq~\ref{eq:surrogate1}, we can get
\begin{equation}
\begin{split}
    \frac{1}{T} \sum_{t=1}^{T} p_{t} \tilde{Rr}_{t} \leq \frac{1}{T} \sum_{t=1}^{T} \tilde{Rr}_{t} + \rho \sqrt{\mathrm{Var}(\tilde{Rr}_{t})} \\
    \Rightarrow \frac{1}{T} \sum_{t=1}^{T} (1-p_{t}) \tilde{Rr}_{t} \geq \rho \sqrt{\mathrm{Var}(\tilde{Rr}_{t})}
\end{split}
\end{equation}
The above formulation indicates that the larger weighted sum of relative rate $\tilde{Rr}_{t}$ means the less variance of $\tilde{Rr}_{t}$. 

Thus, the problem can be re-written as Eq~\ref{eq:bandit_problem}. We can formulate problem of finding $\mathbf{p}$ as an adversarial multi-armed bandit problem in which the player chooses an arm from $\mathcal{P}_{\rho, \mathbf{T}}$ to maximize $\frac{1}{T} \sum_{t=1}^{T} p_{t} \tilde{Rr}_{t}$. 

\begin{equation}
    \max_{\mathbf{p} \in  \mathcal{P}_{\rho, \mathbf{T}}} \sum_{t=1}^{T} p_{t} \frac{\alpha_{t}\mathcal{L}_{t}(\theta)}{||\boldsymbol{g}_{t}||} \label{eq:bandit_problem}
\end{equation} 

To solve above problem, we use Mirror Gradient Ascent method to get $\mathbf{p}$. Following the BanditMTL~\cite{mao2021banditmtl}, we select Legendre Function $\Phi_{p}(p) = \sum_{t=1}^{T} p_{t}\log p_{t}$. We set $p^{0}_{t} = \frac{1}{T}$, and we can get 
\begin{equation}
    p_{t}^{k+1} = \frac{e^{\frac{1}{1+\lambda}(\log p_{t}^{k}+\eta_{p} \tilde{Rr}_{t})}}{\sum_{t=1}^{T} e^{\frac{1}{1+\lambda}(\log p_{t}^{k}+\eta_{p} \tilde{Rr}_{t}})}
\end{equation} where $\eta_{p}$ is the step size for the player, and $\lambda$ is the solution of Lagrangian of  Legendre Function in dual space. Detailed algorithm please refer to Appendix.

Finally, we integrate $\boldsymbol{p}^{k+1} = (p_{1}^{k+1},...,p_{T}^{k+1})$ and GBM generated weights $\boldsymbol{\alpha}$ to form final objective function. As mentioned in last paragraph of section~\ref{sec:theo_analysis}, we only get empirical variance on a mini-batch data and partially minimize this loss using gradient descent. Thus, we adopt the strategy like Reptile~\cite{nicol18reptile} to merge $\boldsymbol{\alpha}$ and $\boldsymbol{p}^{k+1}$. The final objective function of k+1 updating step is as follow:
\begin{equation}
    \min_{\theta} \{\frac{1}{T} \sum_{t=1}^{T} (\epsilon\alpha_{t}\mathcal{L}_{t}(\theta) + (1-\epsilon) p_{t}^{k+1} \frac{\alpha_{t}\mathcal{L}_{t}(\theta)}{||\boldsymbol{g}_{t}||}) \}\label{eq:obj_final}
\end{equation}
If $\epsilon$ is 1, our EMTL deteriorates to the GBM method. The proposed method is presented in algorithmic form in Algorithm~\ref{alg}. We assume the training procedure has K steps.






\begin{algorithm} 
	\caption{EMTL} 
	\label{alg} 
	\begin{algorithmic}
		\REQUIRE  data $\{D_{t}\}_{t=1}^{T}$, the learning rate $\eta$ and hyperparamter $\epsilon$.
		\STATE Initial parameters: $\boldsymbol{p}^{0} = (\frac{1}{T},\frac{1}{T},...,\frac{1}{T})$, randomly initialize $\theta^{0}$
		
		\FOR {$k \leftarrow 1$ to $K$}
            \STATE Compute the gradient of every tasks on shared parameters $\boldsymbol{G} = [\boldsymbol{g}_1,...,\boldsymbol{g}_{T}]$
            \STATE Use a GBM method to get weights $\boldsymbol{\alpha} = \mathrm{GBM}(\boldsymbol{G})$
            \STATE Compute relative rate of loss and gradient norm $\tilde{\boldsymbol{Rr}} = [\alpha_{1}\mathcal{L}_{1}/||\boldsymbol{g}_{1}||,...,\alpha_{T}\mathcal{L}_{T}/||\boldsymbol{g}_{T}||]$
            \STATE Use Mirror Gradient Ascent method on relative rate of loss and gradient norm
            $$
                 p_{t}^{k+1} = \frac{e^{\frac{1}{1+\lambda}(\log p_{t}^{k}+\eta_{p} \tilde{Rr}_{t})}}{\sum_{t=1}^{T} e^{\frac{1}{1+\lambda}(\log p_{t}^{k}+\eta_{p} \tilde{Rr}_{t}})}
            $$
            \STATE \textbf{Update} parameter $\theta$:
            $$
                \theta^{k+1} = \theta^{k} - \eta \nabla_{\theta} \frac{1}{T} \sum_{t=1}^{T} (\epsilon\alpha_{t}\mathcal{L}_{t}(\theta) + (1-\epsilon) p_{t}^{k+1} \frac{\alpha_{t}\mathcal{L}_{t}(\theta)}{||\boldsymbol{g}_{t}||})
            $$
        \ENDFOR
		\RETURN $\theta$ with best validation performance
	\end{algorithmic} 
\end{algorithm}


\subsection{Discussion}\label{sec:discussion}
In this section, we discuss some important differences between EMTL and two previous SOTA methods, IMTL and BanditMTL.

IMTL~\cite{liu2021imtl} first proposes the concept of impartial for MTL. But they straightly force the absolute contribution for updating shared parameters of different tasks to be same. Specifically, IMTL-G forces aggregated gradient of task-shared parameters has equal projections onto the raw gradients of individual tasks, and IMTL-L scales all task-specific losses to be same.  IMTL can not provide theoretical guarantee of converge, so it can settle for a sub-optimal solution for the sake of fairness~\cite{navon2022nashmtl}. Our EMTL considers equity for MTL in a more rational way, including both loss and gradient. Furthermore, we theoretically show that our method can improve generalization of MTL.

BanditMTL~\cite{mao2021banditmtl} first proposes a method to add variance regularization on task loss for MTL. In reality, this method often focuses on the tasks with large loss scale, so it needs to adopt some heuristics to balance loss scale~\cite{9893398}. EMTL uses the same strategy to add variance regularization as BanditMTL, but EMTL achieves much more better performance than BanitMTL. On the one hand, EMTL regularizes tasks' relative rate of loss and gradient norm, rather than task losses. In this way, our method is robust against large loss scale difference. On the other hand, EMTL integrates GBM method and variance regularization, alleviating gradient conflict problem.

\section{Experiments}
\subsection{Experimental Setup}
\subsubsection{Datasets}
We conduct experiments on two different mulit-task research problems, CTR\&CTCVR prediction and scene understanding. We also conduct both offline and online experiments on commercial recommendation system. Tasks of industrial dataset includes classification and regression.

For CTR\&CTCVR prediction, we conduct experiments on both public available benchmark dataset, AliExpress Dataset\footnote{https://tianchi.aliyun.com/dataset/74690}, and industrial datasets. AliExpress dataset collects user logs from real-world traffic in AliExpress e-commercial platform. For offline experiments, we use data from 4 countries, \emph{i.e.} Netherlands, Spain, France and USA. It contains two binary classification tasks, Click Through Rate (CTR) prediction and post-click conversion rate (CTCVR) prediction. 

For scene understanding, we follow the protocal of~\cite{liu2019mtan,navon2022nashmtl} on the NYUv2 dataset~\cite{nyuv2}. It is an indoor scene dataset that consists of 1449 RGBD images and dense per-pixel labeling with 13 classes. We use this dataset as a three tasks learning benchmark for semantic segmentation, depth estimation, and surface normal prediction.

The industrial dataset is collected from our small video platform. It contains three tasks, $i.e$ CTR prediction, watch time prediction and completion rate prediction. Among three tasks, CTR prediction is a classification tasks, while watch time prediction and completion rate prediction are regression task. Completion rate equals to user's watch time divides duration of the small video. The training set contains one week data with about 770 millions samples, and both the validation and testing sets are composed of one day data with about 110 millions samples.

\subsubsection{Implementation Details and Evaluation Metrics}
For two tasks experiment on recommendation datasets AliExpress, we train two popular multi-task recommendation backbones, \emph{i.e.} MMOE~\cite{ma2018mmoe} and PLE~\cite{tang2020ple} on all datasets with two independent tower a the multi-expert layer. MMOE model contain 8 shared single layer MLP as expert network, while PLE model contains 4 shared expert networks and 4 task-specific expert networks for each task. The structure is decided by grid search. Based on the open-source code\footnote{https://github.com/AvivNavon/nash-mtl} of \cite{navon2022nashmtl}, we re-implement Uncertainty, MGDA, NashMTL, BanditMTL and IMTL. We use Adam optimizer~\cite{kingma2014adam} and the learning rate is 1e-3 with weight decay 1e-6. Batch size is 2048. We adopt AUC~\cite{guo17deepfm} as evaluation metrics, which are widely used in CTR and CTCVR prediction. \textbf{AUC} measures the probability of a positive sample ranked higher than a randomly chosen negative one. Besides, we report average AUC of CTR and CTCVR as the overall performance. It should be noted that in the offline CTR or CTCVR prediction, improvement of AUC on the third decimal place can be seen as a significant improvement.

For scene understanding multi-task experiment, we train a Multi-Task Attention Network (MTAN)~\cite{liu2019mtan} which adds an attention mechanism on top of the SegNet architecture~\cite{Badrinarayanan2017SegNet}. 
Each method is trained for 200 epochs with the Adam optimizer too. The initial learning rate is 1e-4 and is halved to 5e-5 after 100 epochs. Batch size is 2. As the three tasks experiment on NYUV2, we follow the settings of CAGrad \cite{liu2021cagrad} and use $\Delta m\%$ as the metric of overall performance. $\boldsymbol{\Delta m}\%$ means the average per-task performance drop of method $m$ relative to the single task learning (STL) baseline. The STL baseline refers to training task-specific SegNet models as in \cite{liu2021cagrad}.

For all experiments, we choose MGDA~\cite{mgda} as GBM method. Hyperparameter $\epsilon$ is decided by grid search in [0,0.5]. Trade-off hyperparameter $\rho$ is grid searched in [0.1,0.3,0.5,0.8,1.0,1.2]. We find the influence of $\eta_{p}$ is small, thus we set $\eta_{p} = 0.5$ for all experiments to save training cost. We use overall performance metrics , \emph{i.e.} average AUC and $\Delta m\%$ for early stopping.


\subsection{Two tasks experiments on Recommendation Dataset}
We first report evaluation results on recommendation dataset, shown in Table~\ref{tab:twotasks}. We report results with best average AUC, and all results are averages over 5 times training with different random seeds. The performance with different hyperparameter $\rho$ please refer to Appendix. It should be noted that, in the offline CTR or CTCVR prediction, increasing of AUC on the third decimal place can be seen as a significant improvement.

\begin{table*}[htp]
\vspace{-2mm}
\caption{Two Tasks Experimental Result on Aliexpress. AUC Values are averages over 5 random seeds and only keep 4 decimal. The best results in each column are bold. The underline result indicates the best result among the models with the same backbone.}
\label{tab:twotasks}
\centering
\scalebox{0.63}{
\begin{tabular}{l|ccc|ccc|ccc|ccc}
\toprule
\multirow{2}{*}{\textbf{Method}}& \multicolumn{3}{c|}{US} &\multicolumn{3}{c|}{ES} & \multicolumn{3}{c|}{FR} & \multicolumn{3}{c}{NL} \\
& CTR & CTCVR & Average & CTR & CTCVR & Average & CTR & CTCVR & Average & CTR & CTCVR & Average\\\hline
STL &0.7058 &0.8637 &0.7848 &0.7252 & 0.8832&0.8042 & 0.7174 &0.8702&0.7938 & 0.7203 & 0.8556 &0.7880\\\hline
\textbf{MTL Model}& & & & & & & & &&&& \\
MMOE&0.7053 &0.8639 &0.7846 &0.7269 & 0.8899& 0.8084& 0.7226&0.8748&0.7987&0.7195&0.7870&0.7533\\
PLE & 0.7092& 0.8699& 0.7896&0.7268 &0.8861 &0.8065 &0.7252 &0.8679&0.7966&\textbf{0.7268}&0.8571&0.7920\\\hline
\textbf{GBM}& & & & & & & &&&&&\\
MMOE+MGDA &0.7060 &0.8703 &0.7882 &0.7294&0.8901 &0.8098 &0.7279 &0.8766&0.8022&0.7231&0.8608&0.7920\\
PLE+MGDA &0.7066&0.8674&0.7870&0.7278 &0.8928 &0.8103 &0.7263 &0.8806&0.8034&0.7249&0.8583&0.7916\\
MMOE+NashMTL &0.7088&0.8723&0.7905&0.7285&0.8957&0.8121&0.7241&0.8777&0.8009&0.7255&0.8616&0.7935\\
PLE+NashMTL &0.7067 &0.8719&0.7894&0.7292&0.8929&0.8111&0.7228&0.8794&0.8011&0.7262&0.8642&0.7952\\ 
MMOE+IMTL-G &0.7026&0.8643&0.7835&0.7275&0.8922&0.8098&0.7226&0.8742&0.7984&0.7237&0.8609&0.7923\\
PLE+IMTL-G &0.7087&0.8707&0.7897&\underline{0.7304}&0.8897&0.8101&0.7255&0.8796&0.8026&0.7256&0.8645&0.7950\\\hline
\textbf{LBM}& & & & &  &&&&& & &\\
MMOE+Uncertainty &0.7050&0.8670&0.7860&0.7286&0.8932&0.8109&0.7248&0.8779&0.8013&0.7245& 0.8614&0.7930\\
PLE+Uncertainty &0.7061&0.8670&0.7866&0.7291&0.8932&0.8112&0.7258&0.8784&0.8021&0.7241& 0.8622&0.7932\\
MMOE+BanditMTL &0.7073& 0.8698&0.7885&0.7292&0.8949&0.8120&\textbf{0.7284}&0.8781&0.8033&0.7252& 0.8610&0.7931\\
PLE+BanditMTL &0.7068& 0.8678& 0.7873&0.7292& 0.8936& 0.8114&0.7264&0.8781&0.8023&0.7241& 0.8605&0.7923\\\hline
\textbf{Ours}& & & & & & &&&&& &\\
MMOE+EMTL & \textbf{0.7140}&\textbf{0.8788} &\textbf{0.7964} &\textbf{0.7311}& \textbf{0.8966}& \textbf{0.8139}& 0.7274&\underline{0.8823}&\underline{0.8048}&\underline{0.7268}&\underline{0.8628}&\underline{0.79478}\\
PLE+EMTL & \underline{0.7111}& \underline{0.8725}& \underline{0.7918}& 0.7302& \underline{0.8945}& \underline{0.8123}&\underline{0.7276}&\textbf{0.8824}&\textbf{0.80512}&0.7266&\textbf{0.8654} &\textbf{0.7960}
\\
\bottomrule
\end{tabular}}
\vspace{-2mm}
\end{table*}

According to the table, we can draw three conclusions: 1) EMTL significantly outperforms STL at all metrics and SOTA MTL baselines on 11 out of 12 metrics. For example, EMTL improves MMOE over 0.01 at CTCVR and average AUC on US dataset. These results indicate the effectiveness of our method. 2) The generality of our method is better than other baselines. EMTL achieves best average AUC on all datasets among same backbone. Meanwhile, other MTL methods can not improve backbone on overall performance on all datasets. 3) EMTL has better robustness against unbalanced loss than other multi-task optimization method. Though, averagely the loss of CTR prediction is about 23 times larger than CTCVR, but our proposed method still stably boost backbones. Meanwhile other scale-free method, like NashMTL and IMTL-G, can not consistently improve backbones. As BanditMTL, we use validation loss to balance loss scale just as suggested by~\cite{9893398}, but BanditMTL achieves much worse performance than ours. This indicates the rationality of our method to regularize relative rate rather than task loss or gradient norm straightly.

\subsection{Three tasks experiments on scene understanding}

The three tasks experiment results on scene understanding are shown in Table~\ref{threetasks}. Based on the source code and pre-processed data provided by NashMTL~\cite{navon2022nashmtl}, we re-implement our EMTL and BanditMTL method on NYUv2. Then we train EMTL and BanditMTL as the same protocol as NashMTL. All results in table is averaged by 3 training with different random seed. For every training, we save the results with best $\Delta m\%$ as the final results. The performance with different hyperparameter $\rho$ please refer to Appendix.

From Table~\ref{threetasks}, we can conclude that: 1) we can see that EMTL outperforms SOTA baselines at 8 out of 10 metrics. Especially, EMTL achieves best performance at overall performance metric $\Delta m\%$ by a significant margin, increasing over 1\% than NashMTL. Besides, EMTL outperforms STL at 8 out of 9 task metrics. Above results indicate the effectiveness of our EMTL on three task learning. 2) Moreover, EMTL achieves better performance than MGDA at surface normal prediction. The inherent biasedness of MGDA~\cite{liu2021imtl} makes it primarily focus on the task with the smallest gradient norm magnitude. Thus, MGDA outperforms other SOTA methods at the tasks significantly. Above results not only indicate the superiority of our method, but also the great robustness against large gradient norm magnitude difference.

\begin{table}[htp]
\vspace{-2mm}
\caption{Three Tasks Experimental Result on NYUv2. Values are averages over 3 random seeds. Arrows indicate the values are the higher the better (↑) or the lower the better (↓). The best results in each column are bold.}\label{threetasks}
\centering
\scalebox{0.8}{
\begin{tabular}{cccccccccccc}\toprule
\multirow{3}{*}{\textbf{Method}} & \multicolumn{2}{c}{\textbf{Segment}}         & \multicolumn{2}{c}{\textbf{Depth}}    &     \multicolumn{5}{c}{\textbf{ Surface Normal }} &\multirow{3}{*}{\textbf{$\Delta$ m}\%$\downarrow$}\\
\cmidrule(r){2-3} \cmidrule(r){4-5} \cmidrule(r){6-10}
             &\multirow{2}{*}{mIoU$\uparrow$}&\multirow{2}{*}{Pix Acc$\uparrow$} & \multirow{2}{*}{Abs Err$\downarrow$}  & \multirow{2}{*}{Rel Err$\downarrow$} &  \multicolumn{2}{c}{Angle Distance$\downarrow$}    &  \multicolumn{3}{c}{Within $t^{\circ}\uparrow$}   &         \\
             \cmidrule(r){6-7} \cmidrule(r){8-10}
             &              &         &          &         & Mean           & Median  & 11.25    & 22.5   & 30      &         \\\hline
STL          & 38.31        & 63.76   & 0.6754   & 0.2780  & 25.01          & 19.21   & 30.15    & 57.20  & 69.18   &        \\\hline
\textbf{LBM} &&&&&&&&&&\\
LS           & 39.29        & 65.33   & 0.5493   & 0.2263  & 28.15          & 23.96   & 22.09    & 47.50  & 61.08   & 5.59   \\
SI           & 38.45        & 64.27   & 0.5354   & 0.2201  & 27.60          & 23.37   & 22.53    & 48.57  & 62.32   & 4.39   \\
RLW          & 37.17        & 63.77   & 0.5759   & 0.2410  & 28.27          & 24.18   & 22.26    & 47.05  & 60.62   & 7.88   \\
DWA          & 39.17        & 65.31   & 0.5510   & 0.2285  & 27.61          & 23.18   & 24.17    & 50.18  & 62.39   & 3.57   \\
Uncertainty  & 36.87        & 63.17   & 0.5446   & 0.2260  & 27.04          & 22.61   & 23.54    & 49.05  & 63.65   & 4.15    \\
BanditMTL    & 39.65        & 65.03   & 0.5959   & 0.2436  & 29.66          & 25.83   & 20.00     & 44.00  & 57.41   & 10.79 \\\hline
\textbf{GBM} &&&&&&&&&&\\
MGDA         & 30.47        & 59.90   & 0.6070   & 0.2555  & 24.98          & 19.56   & 29.09    & 56.74  & 69.21   & 1.38  \\
PCGrad       & 38.04        & 64.64   & 0.5550   & 0.2325  & 27.41          & 22.80   & 23.86    & 49.83  & 63.14   & 3.97   \\
CAGrad       & 39.81        & 65.49   & 0.5486   & 0.2250  & 26.31          & 21.58   & 25.61    & 52.36  & 65.58   & 0.43   \\
GradDrop     & 39.39        & 65.12   & 0.5455   & 0.2279  & 27.48          & 22.96   & 23.38    & 49.44  & 62.87   & 3.66 \\
IMTL-G       & 39.25        & 65.41  & 0.5436   & 0.2256  & 26.02          & 21.19   & 26.2     & 53.13  & 66.24   & -0.69  \\
NashMTL      & 40.07        & \textbf{65.73}   & \textbf{0.5348}   & 0.2179 & 25.26   & 20.08    & 28.4   & 55.47   & 68.15   & -4.04 \\
\hline
EMTL &\textbf{40.23}& 65.53   & 0.5595   & \textbf{0.2155}&\textbf{24.70}& \textbf{19.28} & \textbf{30.04}    & \textbf{57.24}  & \textbf{69.48}   & \textbf{-5.40} \\\bottomrule
\end{tabular}}
\vspace{-2mm}
\end{table}

\subsection{Three Tasks Experiments on Industrial dataset}

\textbf{Offline experiment}
We further conduct an offline experiment to investigate the effectiveness of our method in achieving multi-objects recommendation. We test EMTL on ranking of recommendation, and the backbone is MMOE with three independent tower for each task, respectively. The baseline DCN~\cite{wang2017deep} is a single CTR prediction model, so it can not predict watch time and completion rate.

From Table~\ref{tab:industrial}, we can see that EMTL improves MMOE at all three tasks, whether it is classification or regression. Our method increases 0.012 at CTR AUC from MMOE, and the result is comparable to single CTR prediction model DCN. Besides, EMTL decrease MSE 0.1 and 1.2 of completion rate and watch time, respectively. The improvement can be seen as significant in offline experiment of recommendation. To summarize, above results also show the great generality of our EMTL, because it can improve classification and regression at same time.

\begin{table}[htp]
\vspace{-2mm}
    \caption{Offline experiment on industrial dataset of commercial recommendation system. The metric of CTR is AUC, and metric of Completion Rate and Watch time is MSE. Arrows indicate the values are the higher the better (↑) or the lower the better (↓)}\label{tab:industrial}
    \centering
    \begin{tabular}{c|c|c|c}\toprule
         Method&CTR$\uparrow$&Completion Rate $\downarrow$ & Watch time$\downarrow$ \\\hline
         DCN&0.7504&-&-\\
         MMOE& 0.7489&0.3896&13.5471\\
         MMOE+EMTL& 0.7501&0.3792&12.3436\\
         \bottomrule
    \end{tabular}
\vspace{-2mm}   
\end{table}

\textbf{Online A/B test}
We conduct online A/B test on our small video platform from April 13th to May 10th. We use our EMTL to train a mulit-task ranking model. The tasks are same as offline experiment, and we merge the prediction score of three tasks as ranking score. Results show that watch time of user improves 5\%. This online experiment result demonstrates the effectiveness and practicability of our method in the industrial landscape.

\section{Conclusion}
In this paper, we study the equity problem in MTL. We theoretically prove that variance regularization of relative contribution of all the tasks can improve the generalization performance of MTL. Then we propose a novel and near optimal method \textit{EMTL} to efficiently add variance regularization to MTL. EMTL is robust against unbalanced loss or gradient norm, by balancing gradient and loss at same time. Extensive experiments have been conducted on two different research domains and industrial datasets. The evaluation results show that EMTL outperforms SOTA methods on various benchmarks across multiple domains.

{
\small
\bibliographystyle{acm-reference-format}

}

\end{document}